\documentclass[conference]{IEEEtran}
\IEEEoverridecommandlockouts
\usepackage{cite}
\usepackage{amsmath,amssymb,amsfonts}
\usepackage{algorithmic}
\usepackage{graphicx}
\usepackage{textcomp}
\usepackage{xcolor}
\usepackage{cleveref}
\usepackage{algorithm}
\usepackage{mathtools}
\usepackage{graphicx}
\usepackage{subcaption}
\DeclareMathOperator*{\argmax}{arg\,max}
\DeclareMathOperator*{\argmin}{arg\,min}

\DeclareMathOperator{\MLP}{MLP}
\DeclareMathOperator{\SelfAttention}{SelfAttention}

\def\BibTeX{{\rm B\kern-.05em{\sc i\kern-.025em b}\kern-.08em
    T\kern-.1667em\lower.7ex\hbox{E}\kern-.125emX}}
    
\usepackage{eso-pic}
\newcommand\AtPageUpperMyright[1]{\AtPageUpperLeft{%
 \put(\LenToUnit{0.17\paperwidth},\LenToUnit{-2cm}){%
     \parbox{0.9\textwidth}{\raggedleft\fontsize{8}{11}\selectfont #1}}%
 }}%
\newcommand{\conf}[1]{%
\AddToShipoutPictureBG*{%
\AtPageUpperMyright{#1}
}
}    

\begin{document}


\title{\vspace*{1cm}EqDrive: Efficient Equivariant Motion Forecasting with Multi-Modality for Autonomous Driving \\
}

\author{\IEEEauthorblockN{Yuping Wang}
\IEEEauthorblockA{
\textit{University of Michigan}\\
Ann Arbor, United States \\
ypw@umich.edu}
\and
\IEEEauthorblockN{Jier Chen}
\IEEEauthorblockA{
\textit{Shanghai Jiao Tong University}\\
Shanghai, China \\
adam213@sjtu.edu.cn}
}

\maketitle
\conf{\textit{Preprint}}

\begin{abstract}
Forecasting vehicular motions in autonomous driving requires a deep understanding of agent interactions and the preservation of motion equivariance under Euclidean geometric transformations. Traditional models often lack the sophistication needed to handle the intricate dynamics inherent to autonomous vehicles and the interaction relationships among agents in the scene. As a result, these models have a lower model capacity, which then leads to higher prediction errors and lower training efficiency. In our research, we employ EqMotion, a leading equivariant particle, and human prediction model that also accounts for invariant agent interactions, for the task of multi-agent vehicle motion forecasting. In addition, we use a multi-modal prediction mechanism to account for multiple possible future paths in a probabilistic manner. By leveraging EqMotion, our model achieves state-of-the-art (SOTA) performance with fewer parameters (1.2 million) and a significantly reduced training time (less than 2 hours).
\end{abstract}


\section{Introduction}
For autonomous vehicles, accurately predicting the future trajectories of surrounding vehicles is paramount. While several contemporary models offer promising results in motion forecasting \cite{gao2020vectornet, ma2021continual, girase2021loki, xie2023cognition, chai2019multipath, yuan2021agentformer, li2021spatio, li2021rain, lanegcn, hivt}, many fall short in accounting for the equivariant nature of vehicle trajectories and the invariance between vehicle-vehicle interactions. Recognizing and leveraging the patterns of these can be a significant advantage in predicting the vehicles' future motions. Furthermore, a critical observation underpinning our work is the inherent uncertainties that come with dynamic road environments. Real-world vehicles often exhibit behaviors that can have multiple plausible future paths.

In this paper, we introduce a novel model that seamlessly combines these two crucial aspects: recognizing the equivariant and invariant patterns in vehicle trajectories by employing EqMotion \cite{xu2023eqmotion}, an equivariant
particle and human prediction model with considerations for invariant agent interaction, as the network backbone. In addition, our model predicts multiple future trajectories for each vehicle with corresponding probabilities. By doing so, we not only capture the inherent relations between the vehicle histories but also prioritize their futures based on their likelihood. This multi-modal approach acknowledges the unpredictable nature of road scenarios and provides a richer and more holistic forecast than singular trajectory predictions. Experiments in real-world scenarios show that our model stands out in terms of accuracy and training efficiency.

\section{Related Works}

\subsection{Equivariant Features and Invariant Interactions}
Equivariant feature learning originally rose from the domain of image processing with convolution neural networks (CNN). In \cite{cohen2016group}, the authors proposed a method called Group Equivariant Convolutional Networks that is able to learn image features by preserving their equivariant properties and demonstrated superior image classification performance as compared to conventional CNNs. In the past few years, equivariant feature learning has emerged as a crucial paradigm in machine learning, with widespread applications in areas other than image processing. For example, in \cite{maron2018invariant}, the authors performed a study on recently developed equivariant feature layers and proposed a method that utilizes such layers in graph neural networks. In essence, equivariant representations ensure that transformations in the input lead to equivalent transformations in the output, thus preserving the intrinsic relationships. Such properties are invaluable in computer vision and motion forecasting, where spatial hierarchies and relationships are pivotal. 

On a similar trajectory of thought, Graph Convolution Networks (GCNs) \cite{kipf2016semi} have gained traction due to their capability to capture and model interactions in structured data. GCNs, designed to operate over graph structures, exploit the local symmetries present in data, resulting in invariant representations. These invariants become critical when studying the interactions between nodes, as they ensure that the network's response remains consistent irrespective of the nodes' ordering or specific graph isomorphisms. Combining equivariant feature learning with GCNs paves the way for models that not only recognize and preserve relationships but also ensure consistent interpretations across transformations. This synergy has been explored in a recent work, EqMotion \cite{xu2023eqmotion}, showing promise in complex particle and human body movement prediction tasks that require a deep understanding of interactions and spatial hierarchies. Both equivariant and invariant learning paradigms, when jointly harnessed, can provide a robust foundation for understanding complex structured data. The amalgamation of these techniques offers a promising direction for future research in representation learning.

\subsection{Multi-Modality Motion Forecasting}

Vehicle trajectory forecasting has seen significant advancements in recent years, such as \cite{varadarajan2022multipath++, li2019conditional, li2020evolvegraph, choi2021shared, li2021spatio, sun2022interaction, ma2021multi}. These models employ probabilistic techniques to predict multi-modal futures. In essence, multi-modality allows for the forecasting of multiple potential trajectories for an agent, rather than a single path. To quantify the likelihood of each potential path, models often predict a probability distribution over the set of trajectories. Such probabilistic predictions have been popularized in part by models like the Multiple Futures Prediction \cite{tang2019multiple}. The inherent advantage lies in allowing decision-making systems, like those in autonomous vehicles, to plan and react to the most probable paths while staying aware of less likely but still possible trajectories. Moreover, evaluating the predicted probabilities can provide insights into high-risk or ambiguous scenarios. As systems aim for safer and more informed decisions, multi-modality combined with probabilistic reasoning remains at the forefront of research. This dual-pronged approach ensures that predictions are not only diverse but also anchored in their likelihood of occurrence.

\section{Problem Formulation}
We aim to predict the trajectories of multiple agents based on their past motions and high-definition (HD) maps, while also capturing the inherent uncertainty in their future motions.

Our primary input encompasses the historical trajectories of \(A\) agents. For each agent \(a\) at each time step \(t\) in the input sequence, its location vector is represented as \(x_a^t\) which encompasses its geometric \(x, y\) coordinates. This gives rise to \(A \times T_{\text{in}}\) coordinate pairs, which we symbolize as 
$
  X \in \mathbb{R}^{A \times T_{\text{in}} \times 2}.
$

We also have the HD map as our second input comprised of $L$ centerlines in the scene. Each centerline has $K$ coordinate pairs. We organize all pairs in a matrix $M \in \mathbb{R}^{L \times K \times 2}.$
It is pertinent to note that \( A \), \( H \), \( T_{\text{in}} \), and \( T_{\text{out}} \) are configuration parameters. In conditions where there are fewer than \( A \) agents present, we pad the non-existent values in the matrices with 0 and exclude them from calculations.

Our first output, \( \hat{Y} \), is a compilation of predicted trajectories for each agent. For every agent \(a\), we predict \(H\) potential trajectories over the subsequent \(T_{\text{out}}\) timesteps. Thus, our output has \(A \times H \times T_{\text{out}}\)  coordinate pairs, represented as 
$
    \hat{Y} \in \mathbb{R}^{A \times H \times T_{\text{out}} \times 2}.
$

Additionally, our second output is a set of probabilities correlating with these trajectories, where each entry signifies the probability of the associated trajectory being executed by the agent over all the trajectories. This is denoted by 
$
    P \in \mathbb{R}^{A \times H} \text{ where }  0 <= P_a^h <= 1, 
$ and for each agent the probability of each head sums to 1:
$
    \sum_{h \in H} P_a^h = 1.
$

During the training and validation phases, we access the ground truth future trajectories for all $A$ agents, given as 
$
    Y \in \mathbb{R}^{A \times T_{\text{out}} \times 2}.
$
Our learning objective is to ensure that for each agent the most likely trajectory $ \hat{Y}_a^{\hat{h}}$, where
$
    \hat{h} = \argmax_{h \in H} P_a^h
$
is an accurate reflection of \( Y_a \).

\section{Methodology}

\begin{figure}
    \centering
    \includegraphics[width=\columnwidth,keepaspectratio]{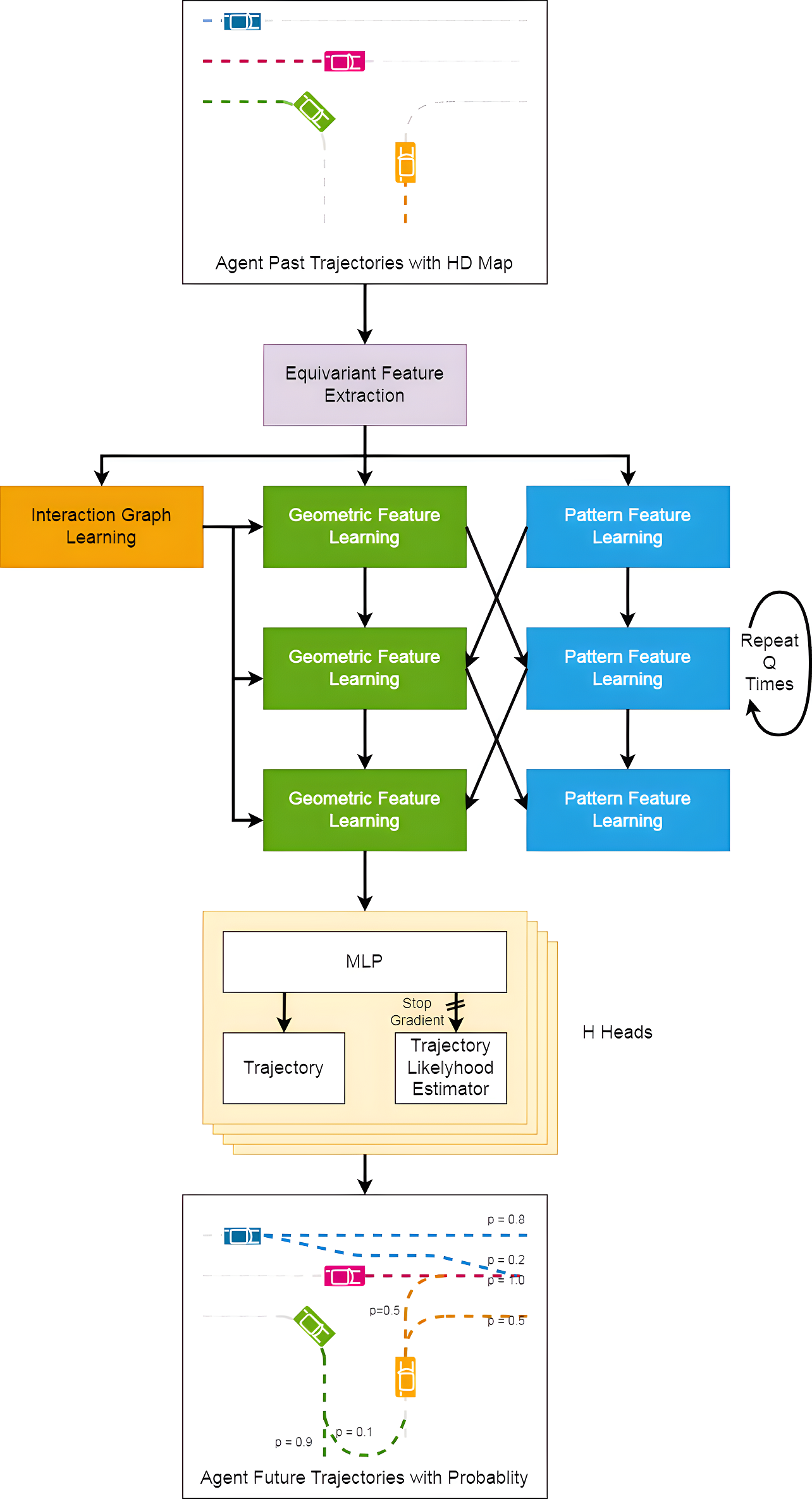}
    \caption{EqDrive Architecture}
    \vspace{-0.5cm}
    \label{fig:model}
\end{figure}

Our model architecture is shown in \Cref{fig:model}. In the following subsections, we explain the model components in detail.

\subsection{Map Information Encoding}

We apply a Self-attention network \cite{vaswani2017attention} to encode the map information into a feature vector.
\begin{equation}
M_\text{encoded} = \SelfAttention(M).
\end{equation}

\subsection{Equivariant Geometry and Invariant Interaction Encoding}

Given the historical trajectories of agents $X$, we first apply the Eqmotion \cite{xu2023eqmotion} as the backbone feature learners. From input $X$, we derive the initial geometric $G^0 \in \mathbb{R}^{A \times T_\text{out}}$ alongside the pattern $H^0 \in \mathbb{R}^{A \times hidden\_dim}$ features as:
\begin{equation}
G^0, H^0 = \mathcal{F}_{\text{InitialFeatureExtractor}}(X).
\end{equation}
The map features is concatenated to the hidden features:
\begin{equation}
H^0 \leftarrow MLP([H^0; M_\text{encoded}]).
\end{equation}
We then model each agent feature as a node and assume a relation between each node. Thus we build a graph convolution mechanism as follows:
\begin{equation}
{e_{ij}} = \mathcal{F}_{\text{GraphConvolution}}(G^0, H^0).
\end{equation}

Based on this, a dual-network structure is consecutively executed, focusing on discerning both the geometric and pattern features. This iterative process takes place $Q$ cycles as shown in \Cref{algo1}.

For an in-depth understanding and implementation of these specific layers, please refer to the EqMotion framework\cite{xu2023eqmotion}. Note that the same operations below are applied to each agent, thus from here on we omit the agent dimension for simplicity.

\begin{algorithm}
\caption{Iterative Process for Feature Learning}
\begin{algorithmic}[1]
\FOR{$q = 1$ \TO $Q$}
    \STATE {\small $G^{q} = \mathcal{F}_{\text{GeometricFeatureExtractor}}(G^{q-1}, H^{q-1}, e_{ij})$} \refstepcounter{equation}\hfill(\theequation)
    \STATE {\small $H^{q} = \mathcal{F}_{\text{PatternFeatureExtractor}}(G^{q-1}, H^{q-1})$} \refstepcounter{equation}\hfill(\theequation)
\ENDFOR
\end{algorithmic}
\label{algo1}
\end{algorithm}

\subsection{Multi-Head Trajectory Decoder and Probability Estimator}

With the final hidden feature vector for each agent $G^q$, we begin the decoding process for $H$ trajectories and a probablistic distribution for them. For each head, we first have an equivariant trajectory decoder implemented with a 4-layer multilayer perception (MLP):
\begin{equation}
    \hat{Y^h} = \MLP(G^P - \Bar{G^P}) + \Bar{G^P}.
\end{equation}
We concatenate the predictions from all heads and employ a final MLP to learn the probability distribution.
\begin{equation}
    \hat{p} = \MLP([\hat{Y^0}, \hat{Y^1};...;\hat{Y^H}]).
\end{equation}

Note that during training, this probability estimator does not pass gradients downstream, because we want the estimator to predict passively based on the learned geometric and pattern features, to avoid interference.

\subsection{Training Time Loss Computation}

During training, we apply a combined loss of a minimum average displacement error (minADE) on the trajectory and a cross entropy loss on the probability. Specifically, given the predictions of all agents at all heads $\hat{Y}$ and ground truth $Y$, we first compute ADE $E$ for each head:
\begin{equation}
    E^h = \frac{1}{T} \sum_{t=1}^{T} \| \hat{Y_t^h} - Y_t \|_2.
\end{equation}
The index to the trajectory with minimum ADE is:
\begin{equation}
    \hat{h} = \argmin_{h \in H} E^h.
\end{equation}
We compute the loss as the ADE on $E^{\hat{h}}$:
\begin{equation}
    L_\text{trajectory} = E^{\hat{h}}.
\end{equation}
We write the ground truth distribution as a one-hot vector:
\begin{equation}
    r \in \mathbb{R}^{H},
    r^h = 
    \begin{cases} 
        0 & \text{for } h \neq \hat{h}, \\
        1 & \text{for } h = \hat{h}.
    \end{cases}
\end{equation}
We then apply a cross-entropy loss on the probability:
\begin{equation}
    L_\text{probability} = - \sum_{h} r^h \log(p^h).
\end{equation}
Lastly, we combine the two losses using a weighted sum controlled by hyperparameter $\beta$, with a higher value in $\beta$ meaning a higher emphasis on trajectory similarity:
\begin{equation}
    L =  \beta L_\text{trajectory} + (1-\beta)L_\text{probability}.
\end{equation}

The above approach ensures that the model not only tries to predict accurate trajectories but also assigns higher probabilities to the most accurate ones. It's a strong combination of losses that encourages both spatial accuracy in predictions and confidence calibration in the predicted probabilities.

\subsection{Inference Time Trajectory Selection}
At inference time, we simply choose the trajectory that has the maximum probability:
\begin{equation}
    \hat{y} = \hat{Y}^{\hat{h}} \text{ and } \hat{h}=\argmax_{h \in H} p^h.
\end{equation}

\subsection{Explanation on the Choice of Loss Function}
Here we want to prove that by applying the cross entropy loss of the probability output, we maximize the likelihood of the true trajectory. Assume our model predicts \( H \) trajectories \( \{\hat{y}^1, \hat{y}^2, \ldots, \hat{y}^H\} \) with associated probabilities \( \{p^1, p^2, \ldots, p^H\} \) and the trajectory closest to the ground truth is indexed at $\hat{h}$. The cross-entropy loss is then:

\begin{equation}
\begin{split}
\text{CrossEntropy} &= - \sum_{h} r^h \log(p^h) \\
                    &= -log(p^{\hat{h}}).
\end{split}                    
\end{equation}

Therefore, by minimizing the cross-entropy loss, we are equivalently maximizing $p^{\hat{h}}$, which is the probability of the trajectory that is closest to the ground truth. This approach ensures that the model gives the highest probability to the trajectory that most closely matches the actual ground truth among the predictions.

\section{Experiments}

\begin{table*}[t!]
\normalsize
\centering
\caption{Comparison of Models}
\label{table:comparison}
\renewcommand{\arraystretch}{1.2}
\begin{tabular}{||p{3cm} | p{2cm} | p{2cm} | p{2cm} | p{2cm} | p{3cm}||} 
\hline
Model & minADE at 3s & minFDE at 3s & Miss Rate & Number of Parameters & Training Time\\
\hline\hline
LaneGCN\cite{lanegcn} & 0.71 & 1.08 & 0.1 & 3.7M & 8hrs w/ 4xTitanX \\\hline
DenseTNT\cite{gu2021densetnt} & 0.75 & 1.05 & 0.1 & 1.1M & 5hrs w/ 8x2080Ti \\\hline
HiVT-128\cite{hivt} & 0.661 & 0.969 & 0.092 & 2.5M & 56.5hrs w/ 1x3060Ti \\\hline
\textbf{EqDrive} & \textbf{0.518} & \textbf{0.915} & \textbf{0.089} & \textbf{1.2M} & \textbf{1.8hrs w/ 1x3060Ti} \\
\hline
\end{tabular}
\vspace{-0.5cm}
\end{table*}

\subsection{Datasets}
We use the Argoverse motion forecasting dataset\cite{chang2019argoverse}. Argoverse offers a training set encompassing approximately 200k scenarios, each detailing the trajectories of both the ego and neighboring agents. The dataset chronicles agent movements from the preceding 2 seconds and forecasts the subsequent 3 seconds, with data points sampled at 10Hz frequency. In addition, map information is also included for all the scenes, represented lane center line waypoints close to the agent.

\subsection{Metrics for Evaluation}
We utilize the following metrics to evaluate the performance of our model: 

\begin{enumerate}
    \item \textbf{Minimum Average Displacement Error (minADE)}: This metric quantifies the average L2 distance between the ground truth trajectory and the closest predicted trajectory. Formally, it is defined as:
    \begin{equation}
        \text{minADE}_{\tau} = \frac{1}{\tau} \min_{h} \sum_{t=1}^{\tau} \lVert \mathbf{y}_{t}^{h} - \mathbf{y}_{t} \rVert_2 
    \end{equation}
    where \(\mathbf{y}_{t}^h\) represents the predicted position at time \(t\) from the \(h\)-th trajectory, \(\mathbf{y}_{t}\) indicates the ground truth position at time \(t\), and \(\tau\) is the prediction horizon. In our case, \(\tau\) can be \(10\) (for 1s), \(20\) (for 2s), or \(30\) (for 3s) given our 10Hz sampling frequency.

    \item \textbf{Minimum Final Displacement Error (minFDE)}: This metric captures the L2 distance between the predicted and actual final positions of the trajectories.
    \begin{equation}
        \text{minFDE}_{\tau} = \min_{h} \lVert \mathbf{y}_{\tau}^h - \mathbf{y}_{\tau} \rVert_2 
    \end{equation}
    Here, \(\tau\) can take values \(10\), \(20\), or \(30\).

    \item \textbf{Miss Rate}: This computes the proportion of predictions that deviate beyond a predefined threshold distance \(d\) from the ground truth.
    \begin{equation}
        \text{Miss Rate} = \frac{1}{N} \sum_{h=1}^{H} \mathbf{1}(\min_{t} \lVert \mathbf{y}_{t}^h - \mathbf{y}_{t} \rVert_2 > d_{threshold})
    \end{equation}
    In the above, \(\mathbf{1}\) is the indicator function which outputs 1 if its internal condition holds true, and 0 otherwise. \(H\) represents the total number of predictions.
\end{enumerate}

\subsection{Implementation Details}

For the training, we use the Adam optimizer with a learning rate of \(10^{-5}\). The model was trained for 50 epochs with a batch size of 512. Our hardware is NVIDIA RTX 3060Ti. Our configuration and hyperparameters are illustrated in the below table:

\begin{center}
\begin{tabular}{||p{4cm} | p{2cm} | p{1cm}||} 
 \hline
Description & Symbol & Value \\
\hline\hline
Length of input sequence & $T_\text{in}$ & 20 \\ \hline
Length of output sequence & $T_\text{out}$ & 30 \\ \hline
Number of total agents in the scene & $A$ & 4 \\\hline
Number of lane centerlines & $L$ & 10 \\ \hline
Number of coordinates per centerline & $K$ & 100 \\ \hline
Number of prediction heads & $H$ & 6 \\\hline
Number of repeats on the feature learning layers & $Q$ & 20 \\ \hline
Size of all hidden dimension & $hidden\_dim$ & 64 \\ \hline
Loss Component Weight & $\beta$ & 0.5 \\
 [1ex] 
 \hline
\end{tabular}
\end{center}

\section{Results and Discussions}

\subsection{Quantitative Results}

We compared our model with current SOTA models including LaneGCN\cite{lanegcn}, DenseTNT\cite{gu2021densetnt} and HiVT\cite{hivt}. EqDrive archives competitive and even slightly better prediction accuracy in terms of minADE, minFDE, and Miss Rate, with comparatively fewer parameters and much shorter training time.

\subsection{Qualitative Results}

\begin{figure*}[!tb]
    \centering
    \begin{subfigure}[b]{0.45\textwidth}
        \centering
        \includegraphics[width=1\linewidth]{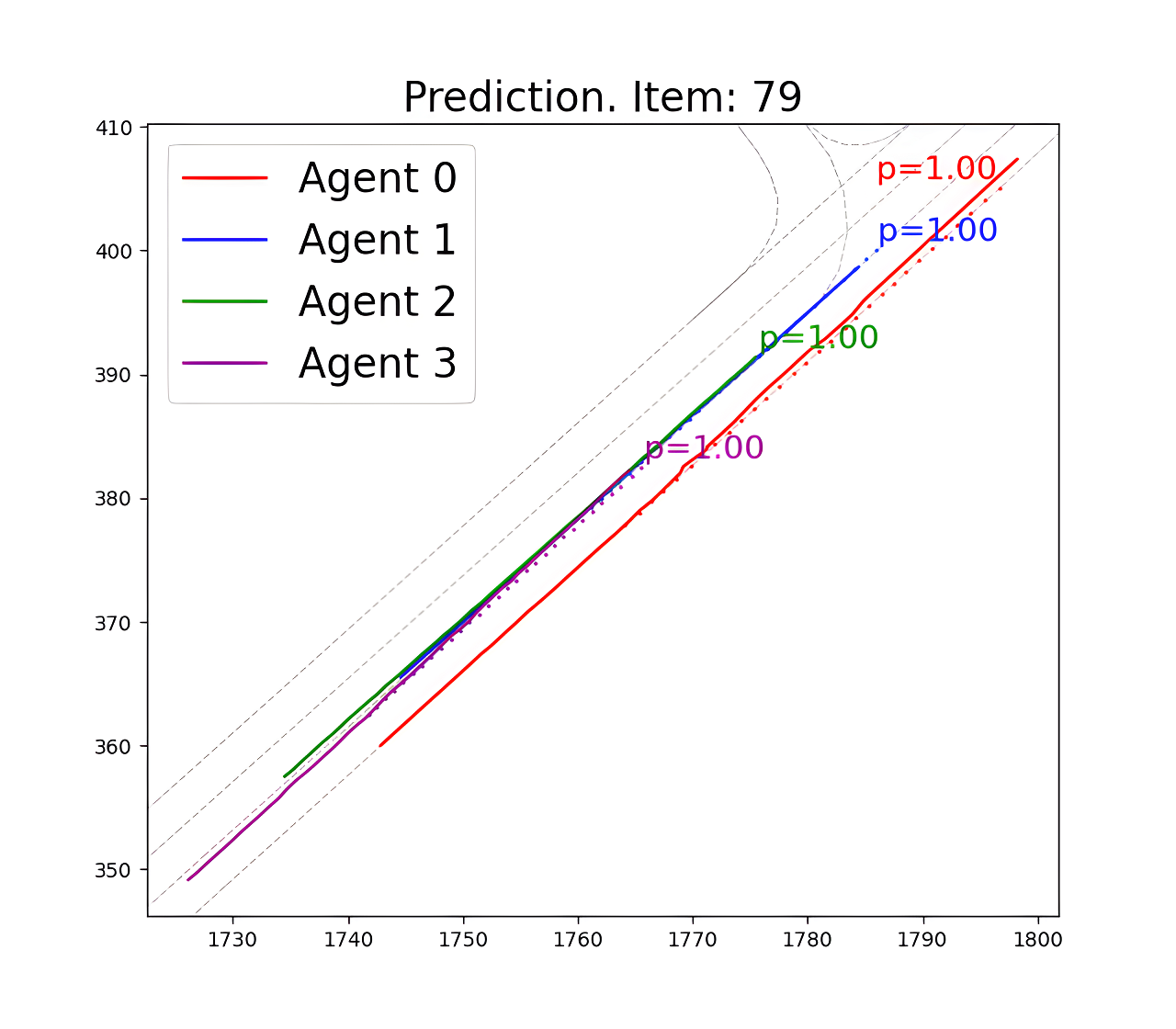}
        \caption{Ego Agent and Neighbors Go Straight}
        \label{fig:straight}
    \end{subfigure}
    \begin{subfigure}[b]{0.45\textwidth}
        \centering
        \includegraphics[width=1\linewidth]{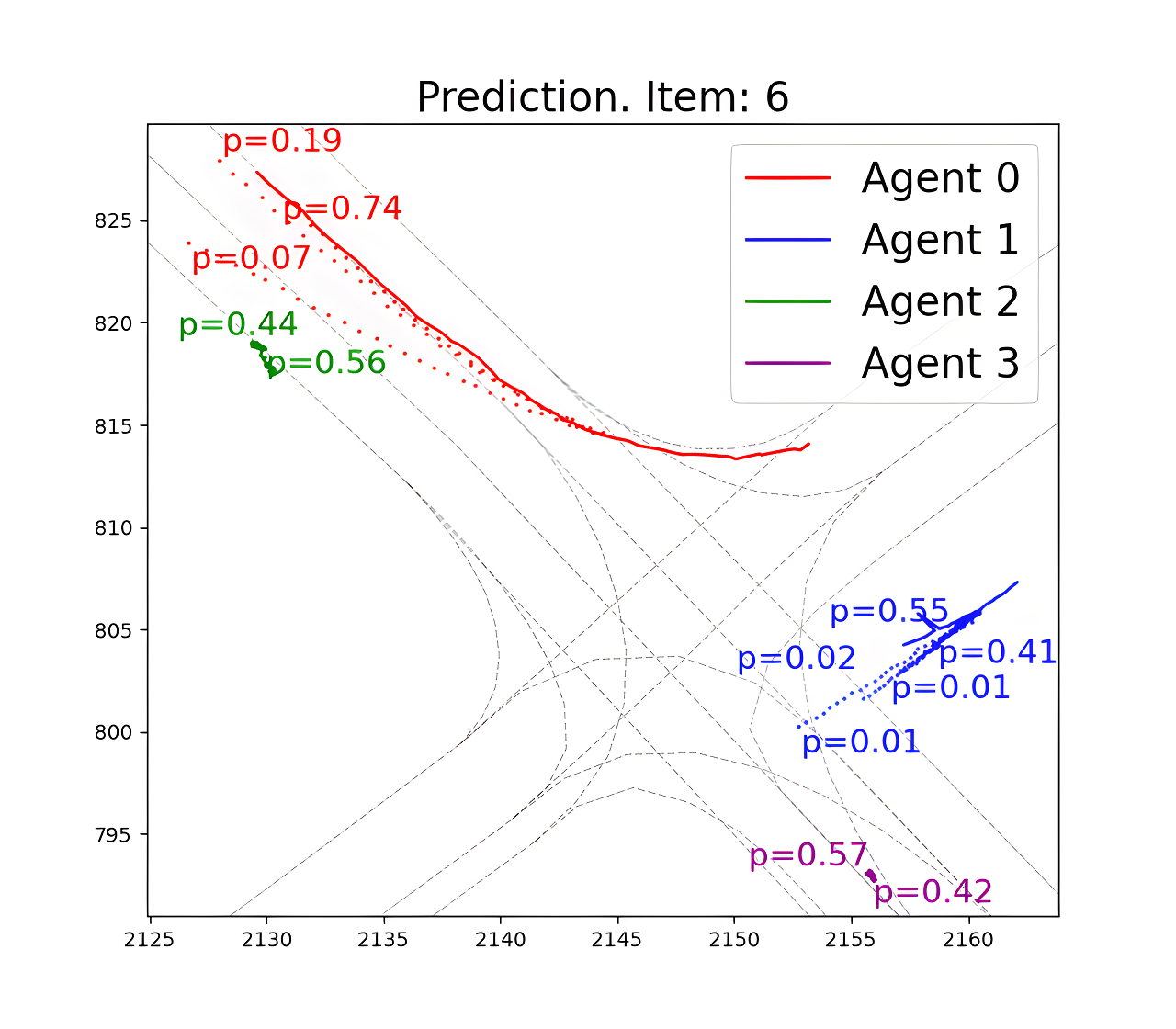}
        \caption{Ego Agent Right Turn}
        \label{fig:right turn}
    \end{subfigure}
    \begin{subfigure}[b]{0.45\textwidth}
        \centering
        \includegraphics[width=1\linewidth]{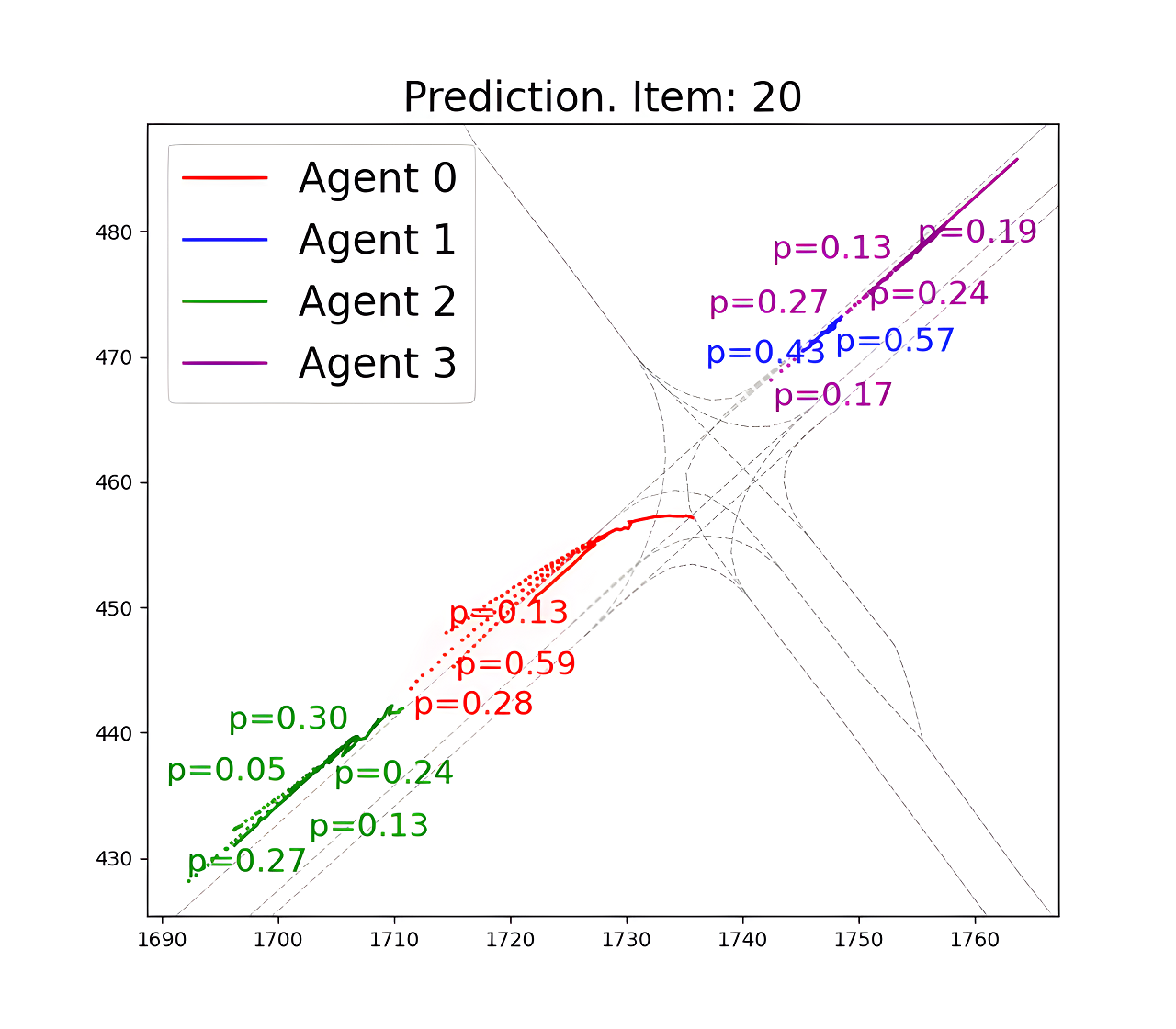}
        \caption{Ego Agent Left Turn}
        \label{fig:left turn}
    \end{subfigure}
    \begin{subfigure}[b]{0.45\textwidth}
        \centering
        \includegraphics[width=1\linewidth]{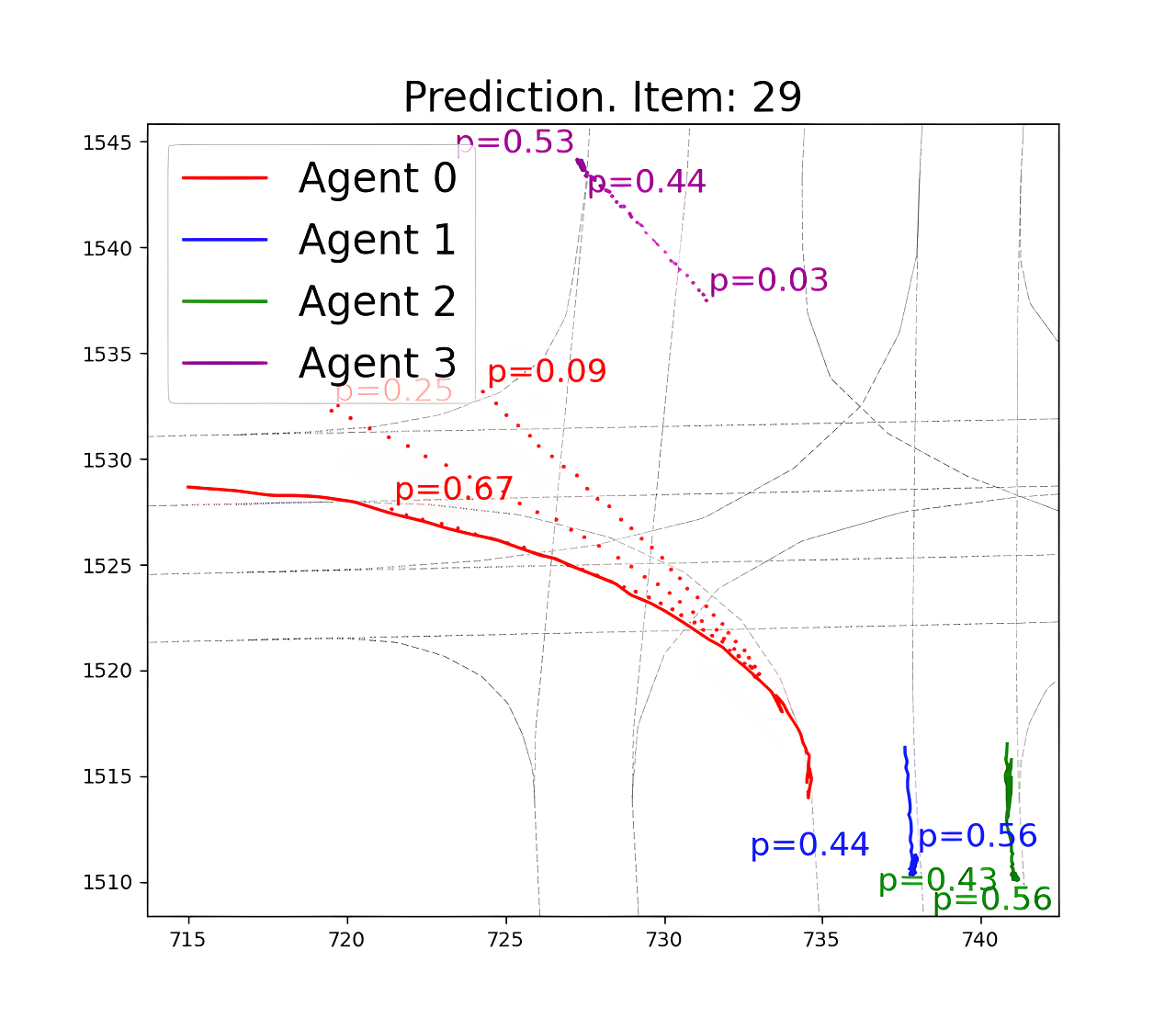}
        \caption{Ego Agent Wide Left Turn}
        \label{fig:wide left turn}
    \end{subfigure}
    \begin{subfigure}[b]{0.45\textwidth}
        \centering
        \includegraphics[width=1\linewidth]{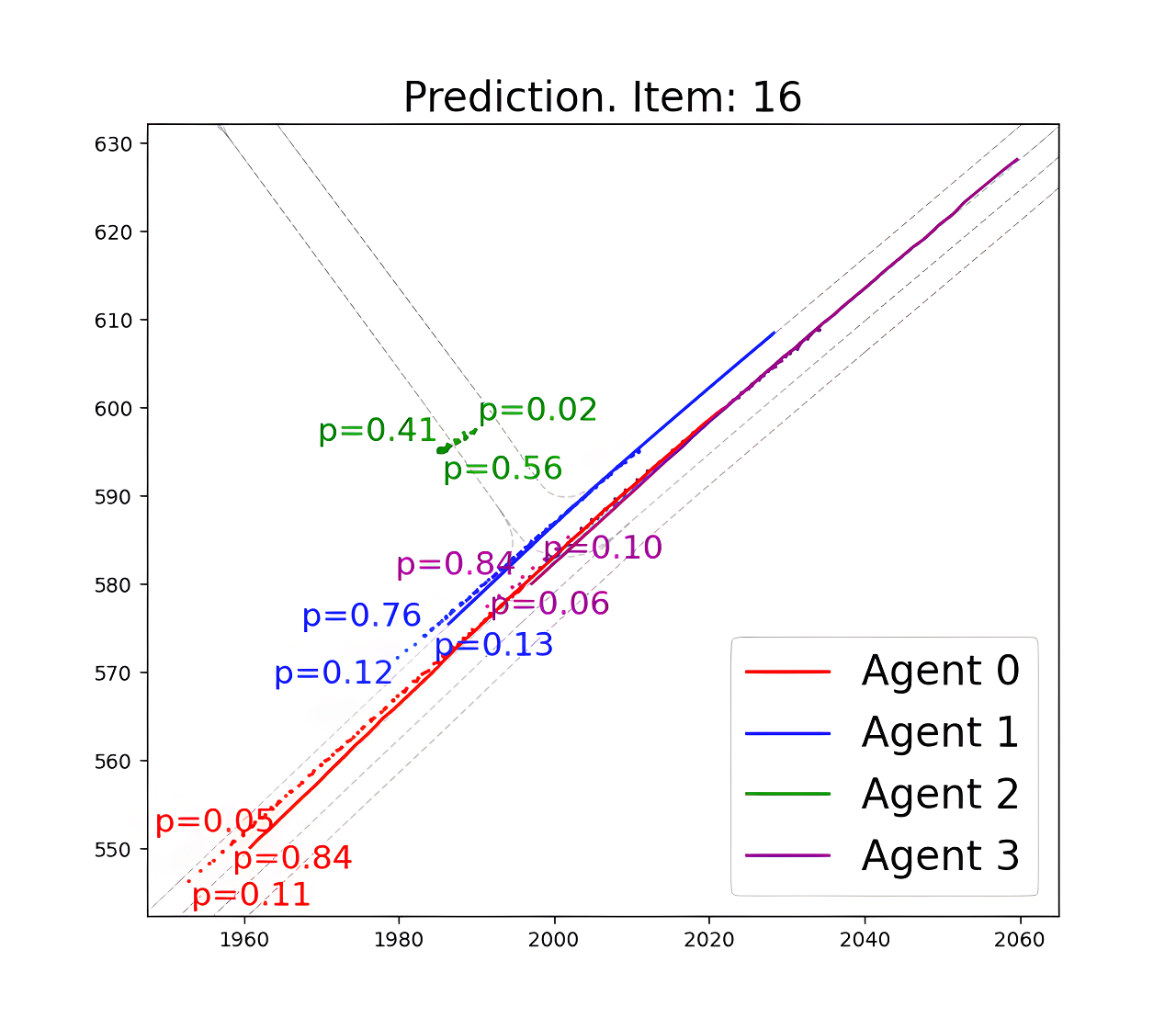}
        \caption{Only Ego Agent Goes Straight}
        \label{fig:ego stright}
    \end{subfigure}
    \begin{subfigure}[b]{0.45\textwidth}
        \centering
        \includegraphics[width=1\linewidth]{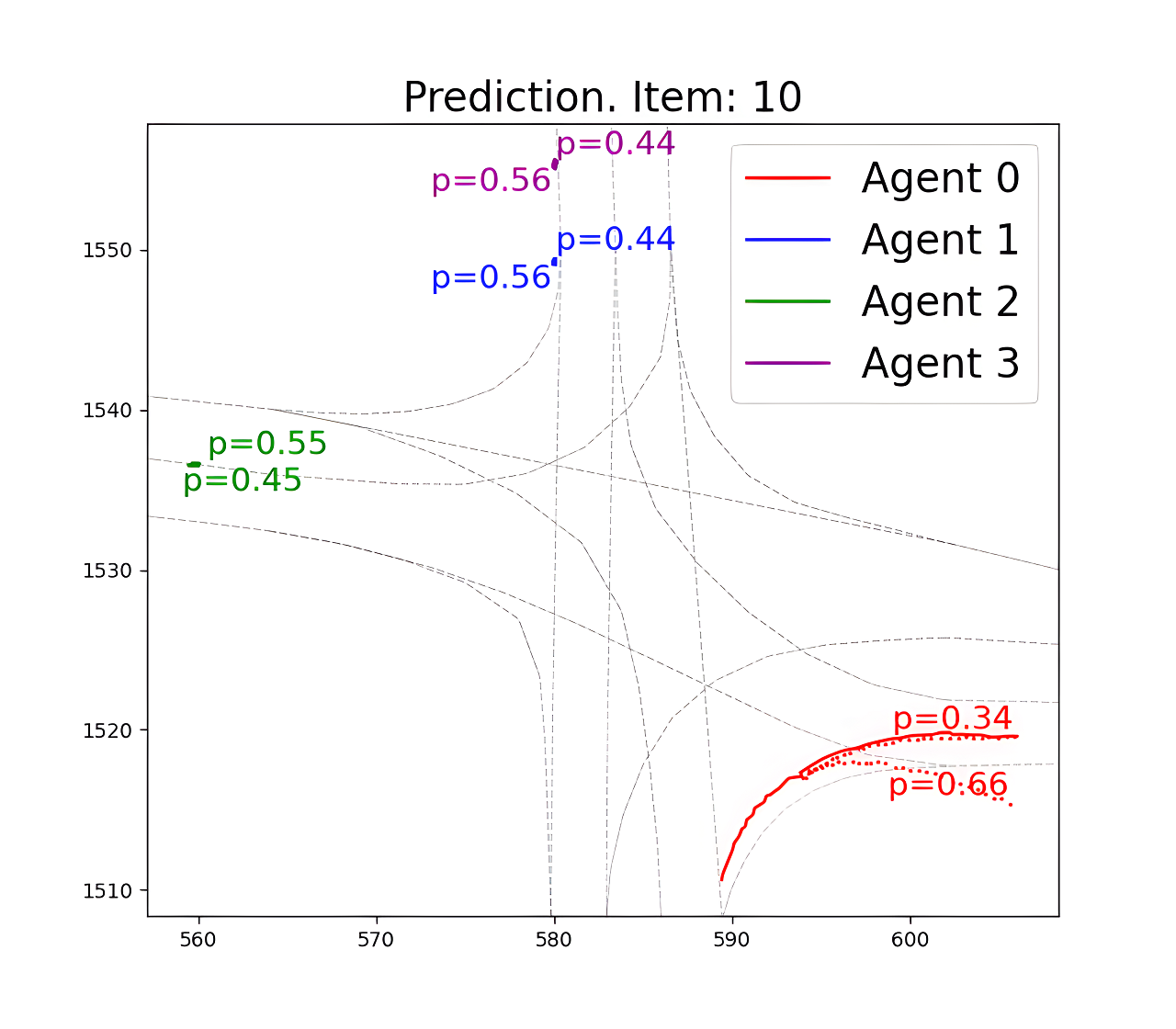}
        \caption{Ego Agent Tight Right Turn}
        \label{fig:tight right}
    \end{subfigure}
    \caption{Visualization of Trajectory Predictions}
    \vspace{-0.4cm}
\end{figure*}

We included \Cref{fig:straight}, \Cref{fig:right turn}, \Cref{fig:left turn}, \Cref{fig:wide left turn}, \Cref{fig:ego stright} and \Cref{fig:tight right} as six visual demonstrations. The solid lines are the 5s historical observation plus ground truth future, while the scatter dots are the 3s prediction. The ego agent is in red. In \Cref{fig:straight} and \Cref{fig:ego stright} we show the simplest scenario where all agents or just the ego agent are predicted to always go straight. The model predicts with high confidence the agents will continue to drive straight, which aligns with the actual on-road situation that we do not vehicles to make sudden turns off their lanes. In \Cref{fig:right turn} and \Cref{fig:tight right}, we see the model provides three predictions that cover the two right turn destination lanes. In \Cref{fig:left turn} and \Cref{fig:wide left turn}, we see the model predicates three possible paths for the ego agent, each corresponding to a different curvature for a left turn. These show our model can capture the realistic vehicular motion where the driver might choose different lanes and/or different turning radius.

In a typical motion planning system for autonomous driving, such as \cite{dolgov2008practical}, the multi-model predictions from our model can provide useful information for planning tasks. For the ego vehicle, the predicted trajectory can provide a high level of heuristics that drives a higher granular free space search algorithm such as A* or rapid exploring random tree (RRT) to produce a drivable path. A realistic guidance makes the search algorithm converge faster \cite{islam2012rrt}. For the neighbor agents, a multi-modal set of predictions allows the planning system to perform probabilistic collision checking  \cite{althoff2009model} and solve for a trajectory that is collision-free at different possible future scenarios.

\section{Conclusion}
By employing EqMotion for autonomous vehicles, we have successfully adopted equivariant models and invariant interactions in autonomous vehicle trajectory prediction. Our introduction of multi-modal predictions with probabilistic heads allows our model to anticipate multiple trajectories, addressing real-world uncertainties. The results, showing we achieve competitive performance with other popular models, validate the effectiveness and efficiency of our approach.

\bibliographystyle{unsrt}
\bibliography{references}

\section{Broader Impact}

This collection of work advances autonomous driving and robotics by improving predictive accuracy, contextual perception, and multi-agent trust. Our contributions to equivariant and multi-modal motion forecasting~\cite{wang2023equivariant, wang2025generative, liu2024toward, wang2025cmp} enable safer and more efficient planning through robust trajectory and occupancy predictions. These models improve downstream decision-making in uncertain, dynamic environments.

In the realm of embodied intelligence, our studies in vision-language alignment and spatial understanding~\cite{xing2025can,xing2025re, xing2025openemma} expose limitations in large vision-language models (LVLMs) for navigation tasks, and introduce retrieval-augmented optimization to mitigate hallucinations and improve grounded reasoning.

Beyond perception and prediction, our protocols for collaboration~\cite{li2025safeflow} and communication~\cite{gao2025airv2x} address safety in multi-agent systems. AirV2X leverages aerial-ground synergy for wider situational awareness, while SafeFlow provides principled mechanisms for secure, concurrent agent interactions. Additionally, our work on novel view synthesis~\cite{hetang2023novel} and point cloud segmentation~\cite{zhang2021point} offers practical benefits in robotic scene understanding and simulation.

We have released key datasets and tools including UniOcc~\cite{wang2025uniocc}, MapBench~\cite{xing2025can}, AirV2X-Perception~\cite{gao2025airv2x}, and SafeFlowBench~\cite{li2025safeflow} to support community adoption. Together, these efforts contribute to more reliable, transparent, and socially beneficial autonomous systems.

\end{document}